\documentclass{article}

\usepackage{microtype}
\usepackage{graphicx}
\usepackage{subfigure}
\usepackage{booktabs} 

\usepackage{hyperref}

\usepackage[accepted]{icml2019}

\usepackage{amsthm}

\newtheorem{Theorem}{Theorem}
\theoremstyle{remark}
\newtheorem{remark}{Remark}

\theoremstyle{definition}
\newtheorem{definition}{Definition}

\usepackage{ enumerate}
\newcommand{\R}{\mathbb{R}}

\newcommand{\E}{\mathbb{E}}

\newcommand{\eps}{\epsilon}

\usepackage{bm}
\newcommand{\N}{\mathcal{N}}

\usepackage{comment}
\usepackage{bm}
\usepackage{amsfonts}
\usepackage{amsmath}
\usepackage{bbm}

\graphicspath{{Figures/}}

\icmltitlerunning{Exponentially-Modified Gaussian Mixture Model}

\begin{document}

\twocolumn[
\icmltitle{Exponentially-Modified Gaussian Mixture Model: \\
        Applications in Spectroscopy}

\begin{icmlauthorlist}
\icmlauthor{Sebastian Ament}{co}
\icmlauthor{John Gregoire}{ct}
\icmlauthor{Carla Gomes}{co}
\end{icmlauthorlist}

\icmlaffiliation{co}{Department of Computer Science, Cornell University, USA}
\icmlaffiliation{ct}{California Institute of Technology, USA}

\icmlcorrespondingauthor{Sebastian Ament}{ament@cs.cornell.edu}

\icmlkeywords{Machine Learning, ICML}

\vskip 0.3in
]

\printAffiliationsAndNotice{}

\begin{abstract}
We propose a novel exponentially-modified Gaussian (EMG) mixture residual model.
The EMG mixture is well suited to model residuals that are contaminated by a distribution with positive support.
This is in contrast to commonly used robust residual models, like the Huber loss or $\ell_1$, which assume a symmetric contaminating distribution and are otherwise asymptotically biased.
We propose an expectation-maximization algorithm to optimize an arbitrary model with respect to the EMG mixture.
We apply the approach to linear regression
and probabilistic matrix factorization (PMF).
We compare against other residual models, including quantile regression.
Our numerical experiments demonstrate the strengths of the EMG mixture on both tasks.
The PMF model arises from considering spectroscopic data.
In particular, we demonstrate the effectiveness of PMF in conjunction with the EMG mixture model on synthetic data and two  real-world applications:   X-ray  diffraction and Raman spectroscopy. We show how our approach is effective in inferring background signals and systematic errors in data arising from these experimental settings, dramatically outperforming existing approaches and revealing the data's physically meaningful components.
\end{abstract}

\begin{figure}
\centering
\includegraphics[width=0.85\linewidth]{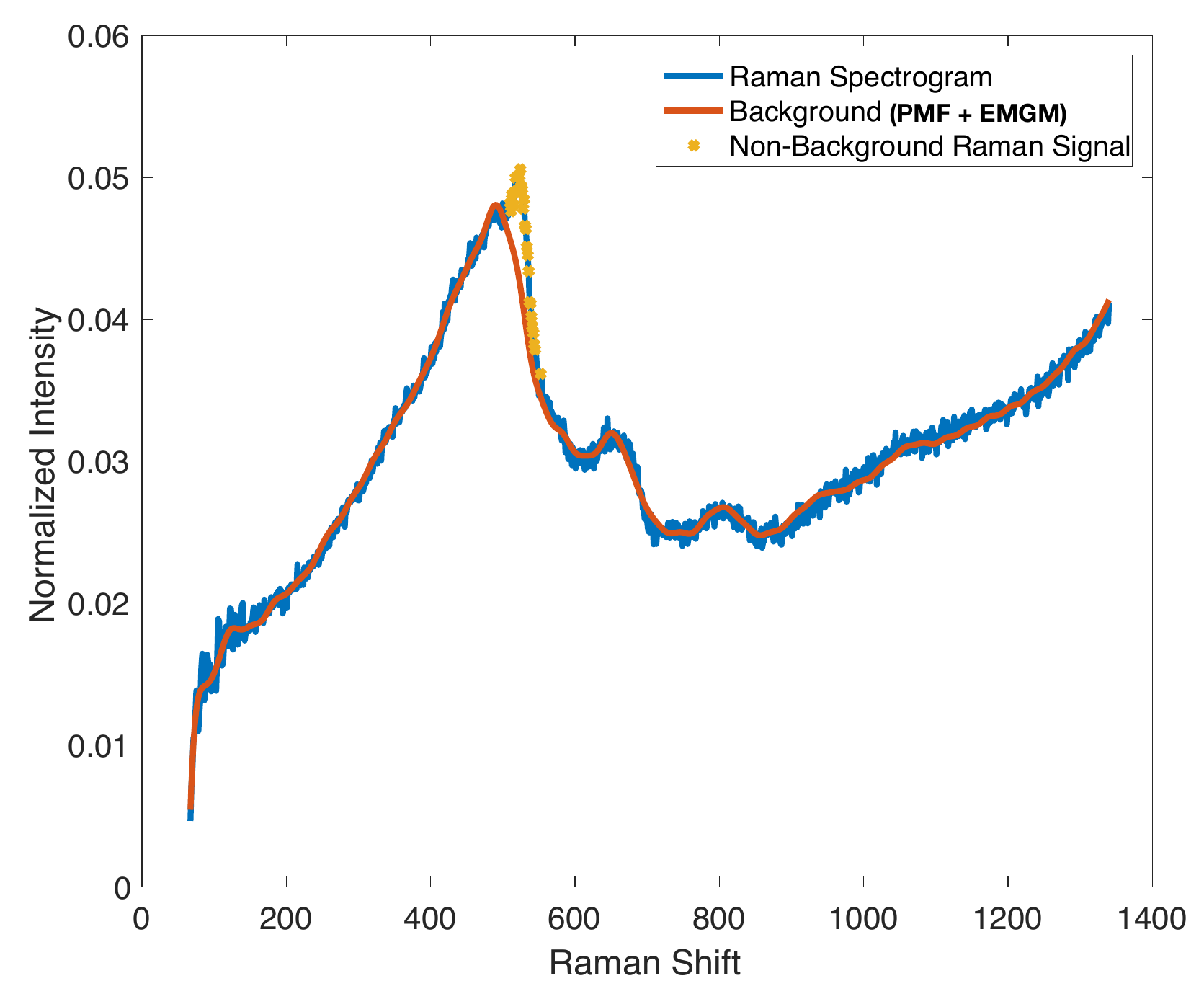}
\caption{ The X axis represents a function of the difference in wavelengths of the incoming and scattered radiation, while
the Y axis represents the intensity of the scattered radiation. 
The figure highlights how a probabilistic matrix factorization (PMF) coupled with the EMG mixture model (in red) 
is able to infer the complex background signal, which defies human ability and existing computational techniques.
}
\label{fig_RamanIntro}
\end{figure}

\section{Introduction}

High-throughput spectroscopic imaging techniques are playing an increasing role in scientific discovery, in areas as diverse as astronomy, biology, materials science, and physics. They also hold promise for commercial applications in a variety of areas such as healthcare, surveillance, consumer products, music, robotics, and autonomous vehicles. As a consequence, we are witnessing an exponential growth in generation rates of spectroscopic data, dramatically  outpacing  humans' ability to analyze them. The grand challenge is therefore to perform  high-throughput unsupervised interpretation of spectroscopic data.

As a motivating example,  consider high-throughput materials discovery in which hundreds or thousands of materials are simultaneously synthesized by deposing a system comprising different chemical elements (typically three or four), onto a substrate \cite{Green2013}. This is analogous to atomic spray painting in which by mixing three or four colors, many new colors are formed.  In order to characterize the crystal structure of the synthesized materials, different spectroscopic imagining techniques such as X-ray diffraction or Raman spectroscopy are used. 
A key challenge is to subtract the X-ray and Raman  patterns of the background substrate material from the X-Ray and Raman patterns of the synthesized materials, due to the fact that the synthesized materials can interact with the background material, compounded with noise in the spectroscopic imaging. 
Furthermore, the background substrate can also exhibit complex patterns as illustrated in  Figure \ref{fig_RamanIntro}. 

These high-throughput experiments often lead to non-negative data. For example, spectroscopic data represent count or intensity quantities, which are naturally non-negative.
To capture the non-negative nature of the data,
we introduce the exponentially-modified Gaussian (EMG) mixture model, which can be applied in arbitrary contexts where residuals are expected to be contaminated by a distribution with positive support.
This is in stark contrast to commonly used robust residual models, like the Huber loss or $\ell_1$, which assume a symmetric contaminating distribution and are otherwise asymptotically biased \cite{Huber1964}.

\textbf{Our contributions:} \textbf{1)} We propose the exponentially-modified Gaussian mixture model, and prove two convexity results for the negative logarithm of its density. \textbf{2)} We further propose an expectation-maximization algorithm to optimize an arbitrary model with respect to the EMG mixture. \textbf{3)}
We contrast the properties of the EMG mixture with commonly-used robust residual models, such as the Huber loss and quantile regression, in a linear regression task. \textbf{4)}
We incorporate the EMG mixture into a probabilistic matrix factorization (PMF) framework, motivated by applications in spectroscopy.
\textbf{5)}
We show the effectiveness of PMF in conjunction with the EMG mixture for the inference of background signals and systematic errors in data arising in X-ray diffraction and Raman spectroscopy.

\section{Preliminaries}

We will denote the normal distribution by $\N(\mu, \sigma)$,
and the normal density evaluated at $x$ as $\N_{\mu,\sigma}(x)$.

\subsection{The Exponentially-Modified Gaussian Distribution}
\label{sec_EMG}

Let the random variable $r$ be defined by
\begin{equation}
\label{eq_EMGRV}
    r = r_E + r_G, \ \ \ r_E \sim \text{Exp}(\lambda), \ \ \ r_G \sim \N(\mu, \sigma).
\end{equation}

The distribution of $r$ is the \textit{exponentially-modified Gaussian distribution} and has previously found applications in biology \cite{EMGBio}, psychology \cite{EMGPsych}, and finance \cite{Carr2008SaddlepointMF}.
Its density is the convolution of an exponential and a Gaussian density
and has the form
\begin{equation}
\label{eq_EMG}
\text{EMG}_{\mu, \sigma, \lambda}(x) = \frac{\lambda}{2} e^{\frac{\lambda}{2} (2\mu+\lambda \sigma^2 - 2x)}
\ \mbox{erfc} \left( \frac{ \mu + \lambda \sigma^2 - x}{\sqrt{2}\sigma} \right),
 \end{equation}
where $\lambda$ is the rate parameter of the exponential random variable,
and $\mu$ and $\sigma$ are the location and scale parameters of the Gaussian random variable, respectively.
Given $\mu = 0$ and a constant $\sigma$, increasing $\lambda$ corresponds to an increased probability of large positive values.

\subsection{Quantile Regression}
\label{sec_quantile}
An important tool of robust statistics is quantile regression.
The $q$th regression quantile is defined as a solution to
\begin{equation}
\label{eq_quantile}
    \min_{b \in \R^n}\left[ \sum_{y_t \geq x_t b} q | y_t - x_t b | + \sum_{y_t < x_t b} (1-q) |y_t - x_t b| \right],
\end{equation}
see \cite{Bassett1978}.
For $q = .5$, this is equivalent to least absolute error regression. 
When the noise distribution
is non-Gaussian, quantile regressions
can be used to build powerful and reliable estimators.
See \cite{koenker2001quantile}
for a discussion including
applications to executive compensation and human birth weights. 

Later, we will estimate quantities related to $\mu$ in \eqref{eq_EMG}.
This is different from the mean of the EMG, which is $\mu + 1/\lambda$.
If the noise were distributed according to \eqref{eq_EMGRV},
and we knew both $\sigma$ and $\lambda$,
we could estimate $\mu$ by computing the quantile with which $\mu$ coincides via the cummulative distribution function and 
\eqref{eq_quantile}.
\textit{The key limitation of this approach for our estimation tasks is that there is no automatic way of choosing the correct quantile if 
the distributional parameters are not known}.

Note also that we can generalize this approach outside of the regression setting, by replacing $x_t b$
in \eqref{eq_quantile} with an arbitrary model.
In section \ref{sec_PMFForSpectroscopy}, we will use this fact to compute a quantile matrix factorization.

\subsection{Probabilistic Matrix Factorization}
\label{sec_ProbMat}
Classical matrix factorization is the problem of finding two matrices $U,V$ such that $A \approx UV$.
The most commonly studied problem is 
\begin{equation}
\label{eq_classicalMF}
\min_{U,V} \|A-UV\|_F^2 = \min_{U,V} \sum_{ij} (A_{ij} - (U V)_{ij})^2.
\end{equation}
Given the inner dimension $k$ of $U$ and $V$,
the solution of this problem can be computed via the singular-value decomposition (SVD) of $A$,
as proven by the Eckart-Young-Mirsky theorem.

\eqref{eq_classicalMF} is equivalent to maximum likelihood estimation of the factor matrices under
a Gaussian error assumption. 
Therefore, an equivalent probabilistic formulation of \eqref{eq_classicalMF} is
\begin{equation}
    \label{eq_probMFLikelihood}
    \max_{U,V} \ \prod_{i,j} \mathcal{N}_{(UV)_{ij},\sigma} (A_{ij})^{I_{ij}},
\end{equation}
where $I_{ij}$ is the indicator function
on the set of indices $\{ij\}$ that are observable.
Importantly, if merely a single entry of $A$ cannot be observed, classical approaches such as SVD
cannot be applied to \eqref{eq_probMFLikelihood}.
\cite{ProbMF} introduced probabilistic matrix factorization in the context of collaborative filtering problems.
In their work, $U$ and $V$ were estimated with maximum a-posteriori (MAP) optimization.

\section{Exponentially-Modified Gaussian Mixture Model}
\label{sec_ResidualModeling}

\subsection{Motivation}

Our approach is motivated by the analysis of spectroscopic data $S$, which can be formally decomposed as
\begin{equation}
   S = P + B,
\end{equation}
where $P$ are spectroscopic peaks, and $B$ are background signals.
Even if the background model were perfect, 
there would be significant differences between the background model and the observed spectrograms; namely $S-B = P$, the non-background spectroscopic peaks.
As we do not know the number of peaks a priori, incorporating peaks explicitly into a model can lead to errors in the analysis whose remedy requires human intervention. 
Our foremost goal is to eliminate the need for human intervention in the analysis of spectroscopic data. 
So instead, we model the residuals caused by peak signals probabilistically.

Spectroscopic data represent counts or intensities and are therefore positive.
The distribution of intensities for a wide range of spectroscopic data can be modeled by the exponential distribution $\text{Exp}(\lambda)$ with rate parameter $\lambda$.
For this reason, the distribution of the residual at the $i$th data point $(S-B)_i$ of a \textit{perfect background fit to noiseless data} can be modeled - again formally - as
\begin{equation}
\label{eq_noiselessDistribution}
    (S-B)_i = P_i \sim p^{\text{noiseless}}_{\lambda, z_i} := (1-z_i) \delta + z_i  \text{Exp}(\lambda),
\end{equation}
where $\delta$ is the Dirac delta distribution and $z_i \in \{0,1\}$.
The variable $z_i$ is $0$ if there is no peak at data point $i$, and $1$ if a peak can be observed.
If the entire data set did not have a single peak, the ideal background model would explain the entire data, so that the residual would be $S-B \sim \delta$.
Similarly, if all data points had observable peaks, $S-B \sim \text{Exp}(\lambda)$.

Data from real experiments is noisy. 
Here, we assume additive Gaussian noise. 
As a result, the residual density is the convolution of the Gaussian density with $p^{\text{noiseless}}_{\lambda, z}$:

\begin{equation}
\label{eq_mainDensity}
\begin{split}
    \text{EMGM}_{\mu, \sigma, \lambda, z}(x) &:= 
    [\mathcal{N}_{\mu,\sigma} \ast p^{\text{noiseless}}_{\lambda, z}](x) \\
    =(1-&z) \N_{\mu, \sigma}(x) + z \ \text{EMG}_{\mu, \sigma, \lambda}(x).
\end{split}
\end{equation}

The EMG density is defined as the convolution of a exponential and a Gaussian density (see section \ref{sec_EMG}).
Thus, for any model $M(\theta)$, with parameters $\theta$, the likelihood of our model with data $D = \{D_i |\  i = 1, \ldots, n \}$ and latent variables $z = \{z_i | \ i = 1, \ldots, n\}$ is
\begin{equation}
\label{eq_likelihood}
\begin{split}
    L(\theta; \sigma, \lambda, z) &= P( D | M(\theta), \sigma, \lambda, z) \\
    &= \prod_{i=1}^n \text{EMGM}_{M_{i}(\theta), \sigma, \lambda, z_{i}}( D_{i} ).
\end{split}
\end{equation}

\eqref{eq_likelihood} is the exponentially-modified Gaussian mixture model we propose for dealing with a contaminating distribution with positive support. 
We introduce the notation $M(\theta)$ to highlight that
an arbitrary model can be optimized with respect to the EMG mixture model.
In our experiments, we let $M$ be a line for linear regression, and also a low-rank matrix 
for the spectroscopic applications.

\begin{remark}
Even though we focus on scientific applications, note that spectroscopic data also abound in other fields like audio source separation, see e.g. \cite{Virtanen2007}.
\end{remark}

\subsection{Theoretical Properties}
\label{sec_theory}
We provide theoretical insights of the mixture by studying the properties of the EMG distribution.
Similar to the Gaussian distribution, the EMG distribution defines a location-scale family.
We define, with a slight abuse of notation, the ``standard" EMG density EMG$_\alpha(x)$ which only depends on one parameter $\alpha$. This simplifies the proofs of the following statements,
minimizes analytical clutter, and
elucidates the function of the individual parameters.

\begin{definition}
Let the standard EMG density be
\begin{equation}
\label{eq_stdEMG}
\text{EMG}_{\alpha}(x) = \frac{\alpha}{2} e^{\alpha (\alpha/2 - x)}
\ \mbox{erfc} \left( \frac{ \alpha - x}{\sqrt{2}} \right).
 \end{equation}
\end{definition}

\begin{Theorem}
Given the standard EMG density \eqref{eq_stdEMG},
we have
\begin{equation}
\label{eq_stdEMGRelation}
    \text{\emph{EMG}}_{\mu, \sigma, \lambda}(x) = \frac{1}{\sigma} \text{\emph{EMG}}_{(\lambda \sigma)}
    \left( \frac{x-\mu}{\sigma} \right).
\end{equation}
\end{Theorem}

\begin{proof}
See supplementary material.
\end{proof}

The negative logarithm of the standard EMG density is 
\begin{equation} 
\begin{split}
- \log \text{EMG}_{\alpha}(x) = - \log \frac{\alpha}{2}           
                        - \alpha \left( \alpha/2 - x \right) \\       
    - \log \text{erfc} \left( \frac{ \alpha - x  }{\sqrt{2}} \right).
\end{split} 
\end{equation}

\begin{Theorem}
\label{lem_convexX}
The negative logarithm of the EMG density is strictly convex in $x$ and $\mu$.
\end{Theorem}

\begin{proof}
See supplementary material.
\end{proof}

\begin{Theorem}
\label{lem_convexL}
The negative logarithm of the EMG density
is strictly convex in $\lambda$ satisfying $1/\lambda > \sigma$.
\end{Theorem}

\begin{proof}
See supplementary material.
\end{proof}

\begin{remark}
The assumption $1/\lambda > \sigma$
implies that the variance of the exponential component is greater than the variance of the Gaussian component. In our applications, this is equivalent to assuming a signal to noise ratio bigger than one.
\end{remark}

Finally, note that the convexity results of Theorem \ref{lem_convexX} and Theorem \ref{lem_convexL} carry over to the negative logarithm of \eqref{eq_likelihood} because sums of convex functions are convex.
Therefore, any non-convexity with respect to the model parameters $\theta$
are strictly due to the model $M(\theta)$ to be optimized,
and not the EMGM.

\subsection{ Expectation-Maximization Algorithm }
\label{sec_EMAlgorithm}

Maximizing the logarithm of \eqref{eq_likelihood} directly is intractable because the discrete variables $z_i$.
Instead, we optimize the log-likelihood of our model using an expectation-maximization algorithm.
To this end, we need to define $\eps := P(z_i = 1)$
and $\gamma_i := \E_{z_i | \theta, \sigma, \lambda, \eps} [z_i]$.
We can then take the expectation of \eqref{eq_likelihood} over all  $z_i$:
\begin{equation}
\label{eq_exptectedLikelihood}
\begin{split}
     \E_{z|\theta, \sigma, \lambda, \eps} ( \log L ) &= \sum_{i=1}^n (1-\gamma_i) \log \N_{M_i(\theta), \sigma}(D_i) \\
     &+ \gamma_i \log \text{EMG}_{M_i(\theta), \sigma, \lambda}(D_i).
\end{split}
\end{equation}

The expectation step is 
\begin{equation}
\begin{split}
    \gamma_i = \frac{\eps \text{EMG}_{M_i(\theta), \sigma, \lambda}(D_i)}{(1-\eps) \mathcal{N}_{M_i(\theta),\sigma}(D_i) + \eps \text{EMG}_{M_i(\theta), \sigma, \lambda}(D_i)}
\end{split}
\end{equation}

The maximization step optimizes \eqref{eq_exptectedLikelihood} for the model parameters $\theta$, the continuous mixture model parameters $\sigma$ and $\lambda$,
and updates the mixture probability $\eps$.
The maximization works in two steps. First, $\sigma, \lambda$ are held fixed while $\theta$ is optimized.
Then, $\theta$ is held fixed while $\sigma, \lambda$ are optimized.
Recall that Theorem \ref{lem_convexL} gives a condition
for when the optimization of $\lambda$ is a strictly convex problem.
If the assumption of the theorem does not hold, we can only guarantee finding a local optimum of this optimization problem.

We use a gradient descent on \eqref{eq_exptectedLikelihood} for both subproblems of the maximization step.
This is not guaranteed to find the true maximum with respect to all unobserved variables.
However, it is guaranteed to improve the likelihood if the parameters do not already constitute a stationary point.
This is a type of generalized EM (GEM) algorithm
and is guaranteed to improve the true likelihood at each iteration until a stationary point is found \cite{ Dempster19977, Neal:1999:VEA:308574.308679}.

The step size of the gradient descent algorithm is chosen using a backtracking line search to ensure descent at every iteration of all optimization procedures in the M-step. 
We re-scale all gradients by the absolute value of their second derivatives, which
makes the descent algorithm scale invariant
and accelerates it in practice \cite{Bertsekas2008}.
All gradient evaluations are linear in the number of parameters. As the EMGM is twice differentiable and strongly convex in $x$ and $\mu$ (see Theorem \ref{lem_convexX}), the number of gradient evaluations scales logarithmically with the required precision.

\section{Related Work}

Herein we  describe prior work in robust statistics, highlighting commonalities and important differences with respect to our work.

Optimization of a model with the likelihood given by \eqref{eq_likelihood}
can be viewed as a location estimation problem of an asymmetrically-contaminated Gaussian distribution.
In particular, we want to estimate $\mu$ for the distribution $F := (1-\eps) \mathcal{N}(\mu,\sigma) + \epsilon C$, where $C$ is a contaminating distribution.

This is related to work in robust statistics starting with Huber's seminal paper \cite{Huber1964},
in which he introduced the function
\begin{equation}
\label{eq_Huber}
    \rho_\delta(x) = \begin{cases}
    \frac{1}{2}x^2 & \text{if } |x| \leq \delta \\
    \delta (x - \frac{1}{2} \delta) & |x| > \delta
    \end{cases}.
\end{equation}

Huber proved that the minimum of $\sum_k \rho(x_k - \xi)$
over $\xi$ is an optimal estimator of the population mean of a contaminated normal distribution.
In particular, he proved that this estimator achieves the minimum asymptotic variance among all translation invariant estimators on
contaminated normal distributions of the form $F = (1-\eps) G + \eps H$, where $G$ is the normal distribution, and $H$ is a contaminating distribution. 
Critically, this optimality result was derived with the assumption of a symmetric contaminating distribution $H$. 
The estimator is not consistent if the contaminating distribution is asymmetric.
Therefore, it is not guaranteed to work well in our setting.

In recent and related work, \cite{Fujisawa2008, Kanamori2015} proposed using scoring rules to guard regression algorithms against substantial contamination. 
Remarkably, these works make no explicit assumptions about the family of contaminating distributions.
However, the methods rely on the $L_2$ inner product of the contaminating distribution and the regular noise distribution to be ``extremely small" \cite{Kanamori2015}.
In other words, all contaminated datapoints have to be exceedingly unlikely under
the regular noise assumption.
This is undoubtedly not true in our scenario \eqref{eq_mainDensity}:
The inner product of the Gaussian and the exponential is equal to EMG$_{0, \sigma, \lambda}(0)$ which is not small in general, except for extremely small $\lambda$.

\cite{Takeuchi2002} introduced the Robust Regression for Asymmetric Tails (RRAT) algorithm.
The algorithm 
can be used to estimate the conditional mean $\mathbb{E}(y | x)$ in a regression setting, and
is based on quantile regression (see Section \ref{sec_quantile}).
It uses the fact that even with an asymmetric contaminating distribution,
there is a quantile which coincides with the mean.
An advantage of RRAT is that it can deal even with heavy-tailed asymmetric contamination, 
as long as its first moment is defined.
However,  this approach has two limitations.
First, though the noise distribution
can be asymmetric,
the algorithm assumes a zero mean.
This is not necessarily true in the case of the EMGM model.
Secondly, even if the approach could be adapted to this setting,
its key limitation
is that its hyper-parameters cannot be chosen automatically.

In contrast to the previously described methods, \textit{all parameters of our model can be automatically inferred from data by optimizing the likelihood with the EM algorithm (see Section \ref{sec_EMAlgorithm}).
This is key for automating scientific discovery in high-throughput settings.}

\section{Experiments}
\label{sec_Experiments}
\begin{figure}
\begin{center}
\includegraphics[width=.8\linewidth]{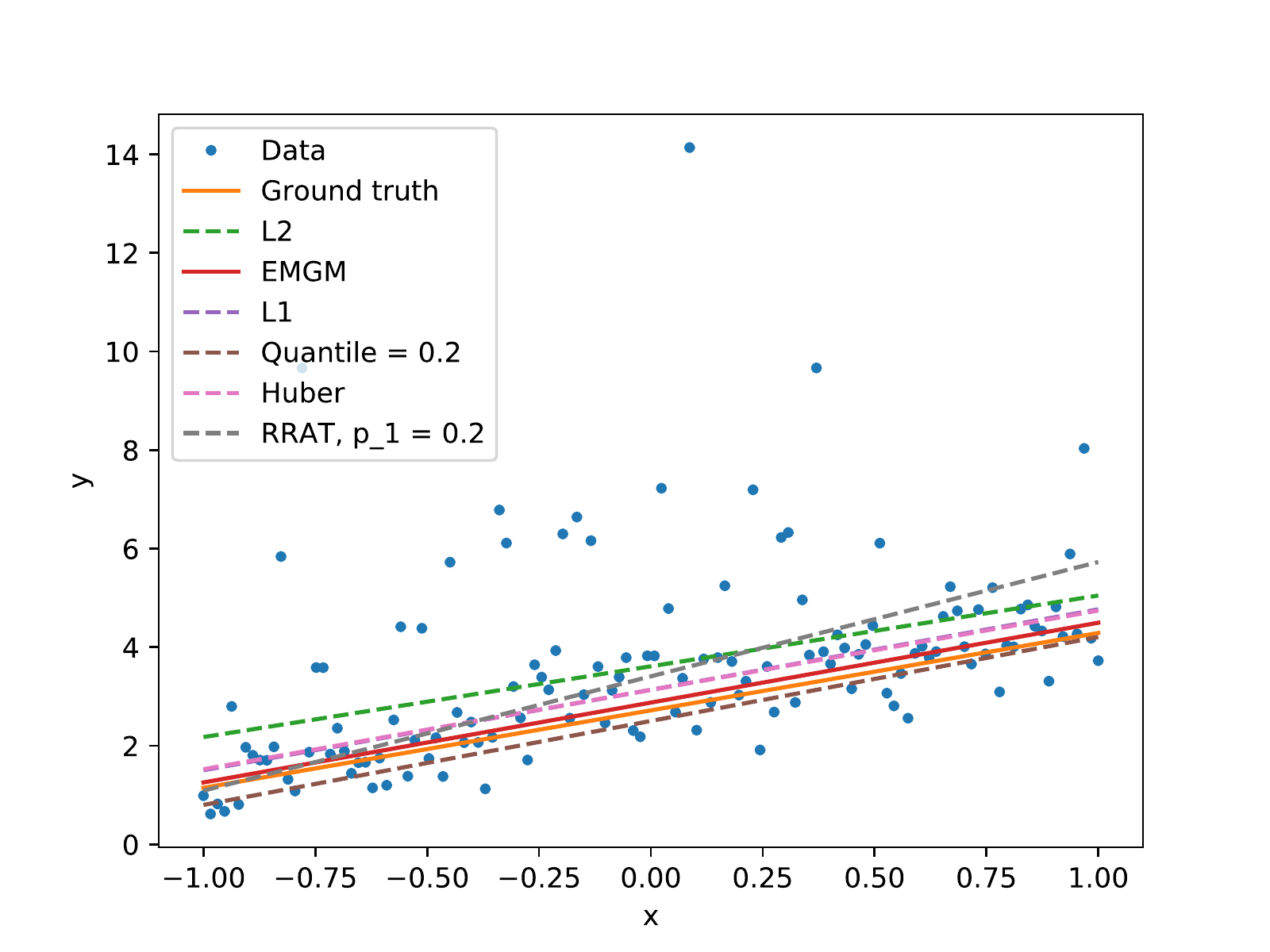}
\caption{A sample dataset with exponentially-distributed contamination and regression results for several residual models.
Most competing methods exhibit positive bias,
while the EMGM is close to the ground truth.
}

\label{fig_linearRegression}
\end{center}
\end{figure}

\subsection{Linear Regression}

We first study the behavior of the EMG mixture residual model on a linear regression task.
Specifically, we are given data points $x, y \in \R$
and want to infer $a,b$ so that $y = ax+b$.
In contrast to the traditional setting,
$y$ is not only corrupted by Gaussian noise, but also by a contaminating distribution with positive support.
In addition to exponential contamination,
we provide results with log-normal contamination to study the behavior of the EMGM algorithm if its distributional assumptions are not satisfied.
In both cases, we let 
\begin{equation}
\label{eq_regressionData}
    y_i = \frac{\pi}{2} x_i + e + G_{i} + \mathbbm{1}_C(i) C_{i},
\end{equation}
where $G_{i} \sim \N(0, 1/2)$, $\mathbbm{1}_C$
is the indicator on the set of contaminated indices,
and $C_{i}$ is drawn from the contaminating distribution.
For all experiments and a given data size $N$, we contaminate 25\% of all data points.
The regression coefficients are initialized to $a = 1, b = 0$.
The initial mixture probability of the EMGM is set to 50\%, and its initial parameters are $\mu = 0, \sigma = 1$, and $\lambda = 1$.

In Figure \ref{fig_linearRegression},
we show regression results on a sample dataset generated using \eqref{eq_regressionData}.
Please see \cite{Takeuchi2002} for details on RRAT.
Notably, $\ell 2$ and RRAT perform worst.
The former is not robust to outliers
and the data breaks RRAT's assumption of zero mean noise.
Even tinkering with the number and type of quantiles
as input to RRAT could not improve its performance on this data.
Therefore we left it out of the evaluations in this paper.
The second tier of residual models are $\ell 1$ and the Huber loss with $\delta = .2$, which visually overlap in the figure.
Though robust against symmetric contamination, these models exhibit positive bias on this data.
Lastly, quantile regression with $q=.2$ and EMGM are visually closest to the ground truth.

\begin{table}
 \caption{ Error statistics of estimation of $a, b$ with different contaminating distributions. MAE is mean absolute value, mean is the mean error, std is the 
 standard deviation of the error. Bold is best.}
 \small
    \centering
\resizebox{.95\linewidth}{!}{
        \begin{tabular}{|c|c|c|c|c|c|c| }
    \multicolumn{5}{c}{Exponentially-Distributed Contamination} \\
    \hline
      $N = 2^8$ & $\ell2$ & Huber & $\ell1$ & Quant .2 & EMGM  \\
    \hline
    MAE a & 1.23e-01 & 6.15e-02 & 6.74e-02 & 6.25e-02 & \textbf{5.08e-02} \\
    mean a & -2.96e-03 & -6.65e-03 & -5.76e-03 & 7.15e-03 & \textbf{-2.93e-03} \\
    std a & 1.58e-01 & 7.65e-02 & 8.23e-02 & 7.95e-02 & \textbf{6.50e-02} \\
    MAE b & 5.02e-01 & 1.63e-01 & 1.59e-01 & 3.27e-01 & \textbf{3.37e-02} \\
    mean b & -5.02e-01 & -1.63e-01 & -1.59e-01 & 3.27e-01 & \textbf{-4.63e-04} \\
    std b & 7.25e-02 & 4.50e-02 & 4.89e-02 & 4.85e-02 & \textbf{4.18e-02} \\
    \hline 
      $N = 2^{14}$ & $\ell2$ & Huber & $\ell1$ & Quant .2 & EMGM  \\
    \hline
    MAE a & 1.49e-02 & 7.90e-03 & 8.32e-03 & 8.56e-03 & \textbf{6.90e-03} \\
    mean $a$ & -8.42e-04 & 8.62e-05 & \textbf{2.05e-04} & 2.38e-04 & 2.93e-04 \\
    std $a$ & 1.84e-02 & 9.96e-03 & 1.03e-02 & 1.07e-02 & \textbf{8.80e-03} \\
    MAE b & 5.01e-01 & 1.64e-01 & 1.61e-01 & 3.28e-01 & \textbf{3.80e-03} \\
    mean $b$ & -5.01e-01 & -1.64e-01 & -1.61e-01 & 3.28e-01 &\textbf{-4.22e-04} \\
    std $b$ & 8.46e-03 & 5.07e-03 & 5.45e-03 & 5.42e-03 & \textbf{4.70e-03} \\
     \hline
     \multicolumn{5}{c}{Log-Normally Distributed Contamination} \\
     \hline
     $N = 2^8$ & $\ell2$ & Huber & $\ell1$ & Quant .2 & EMGM  \\ 
    \hline
    MAE a & 1.20e-01 & 6.33e-02 & 7.04e-02 & 6.73e-02 & \textbf{5.67e-02} \\
    mean a & 6.71e-03 & -7.98e-03 & -8.27e-03 & \textbf{-1.04e-03} & -5.29e-03 \\
    std a & 1.56e-01 & 7.87e-02 & 8.60e-02 & 8.29e-02 & \textbf{7.12e-02} \\
    MAE b & 4.18e-01 & 1.59e-01 & 1.56e-01 & 3.26e-01 & \textbf{3.98e-02} \\
    mean b & -4.18e-01 & -1.59e-01 & -1.56e-01 & 3.26e-01 & \textbf{-2.95e-02} \\
    std b & 7.61e-02 & \textbf{3.89e-02} & 4.22e-02 & 4.69e-02 & 4.13e-02 \\
    \hline 
    $N = 2^{14}$ & $\ell2$  & Huber & $\ell1$ & Quant .2 & EMGM  \\ 
    \hline
    MAE a & 1.50e-02 & 7.87e-03 & 8.70e-03 & 8.27e-03 & \textbf{6.74e-03} \\
    mean a & -1.90e-03 & 7.27e-05 & \textbf{2.60e-05} & -6.46e-04 & -2.96e-04 \\
    std a & 1.84e-02 & 9.64e-03 & 1.05e-02 & 1.02e-02 & \textbf{8.46e-03} \\
    MAE b & 4.13e-01 & 1.59e-01 & 1.57e-01 & 3.28e-01 & \textbf{1.87e-02} \\
    mean b & -4.13e-01 & -1.59e-01 & -1.57e-01 & 3.28e-01 & \textbf{-1.87e-02} \\
    std b & 8.93e-03 & 5.02e-03 & 5.39e-03 & 6.33e-03 & \textbf{4.99e-03} \\
    \hline
    \end{tabular}
}
\label{tab_regressionResults}
\end{table}

For a quantitative comparison of the methods, consider Table \ref{tab_regressionResults}.
We compute the mean absolute error (MAE), the mean error (mean), and the  standard deviation of the error (std) of the estimation of both $a$ and $b$, by considering the regression results of 256 realizations of \eqref{eq_regressionData}
for two data sizes $N$.
All methods seem to be able to estimate $a$ well.
The bias of most methods comes to light when considering $b$.
Indeed, $\ell2$, $\ell1$, and Huber have a negative mean error, indicating positive bias, across both dataset sizes and both contaminating distributions.
The $20\%$ quantile regression has negative bias, 
which is also not mitigated by more data.
In contrast,
the EMGM exhibits small bias and mean absolute error for the exponentially contaminated datasets.
Further, the MAE reduces as the data size increases, indicating that the EMGM correctly adapts its distributional parameters to the data.

\begin{figure}
\begin{center}
\includegraphics[width=.49\linewidth]{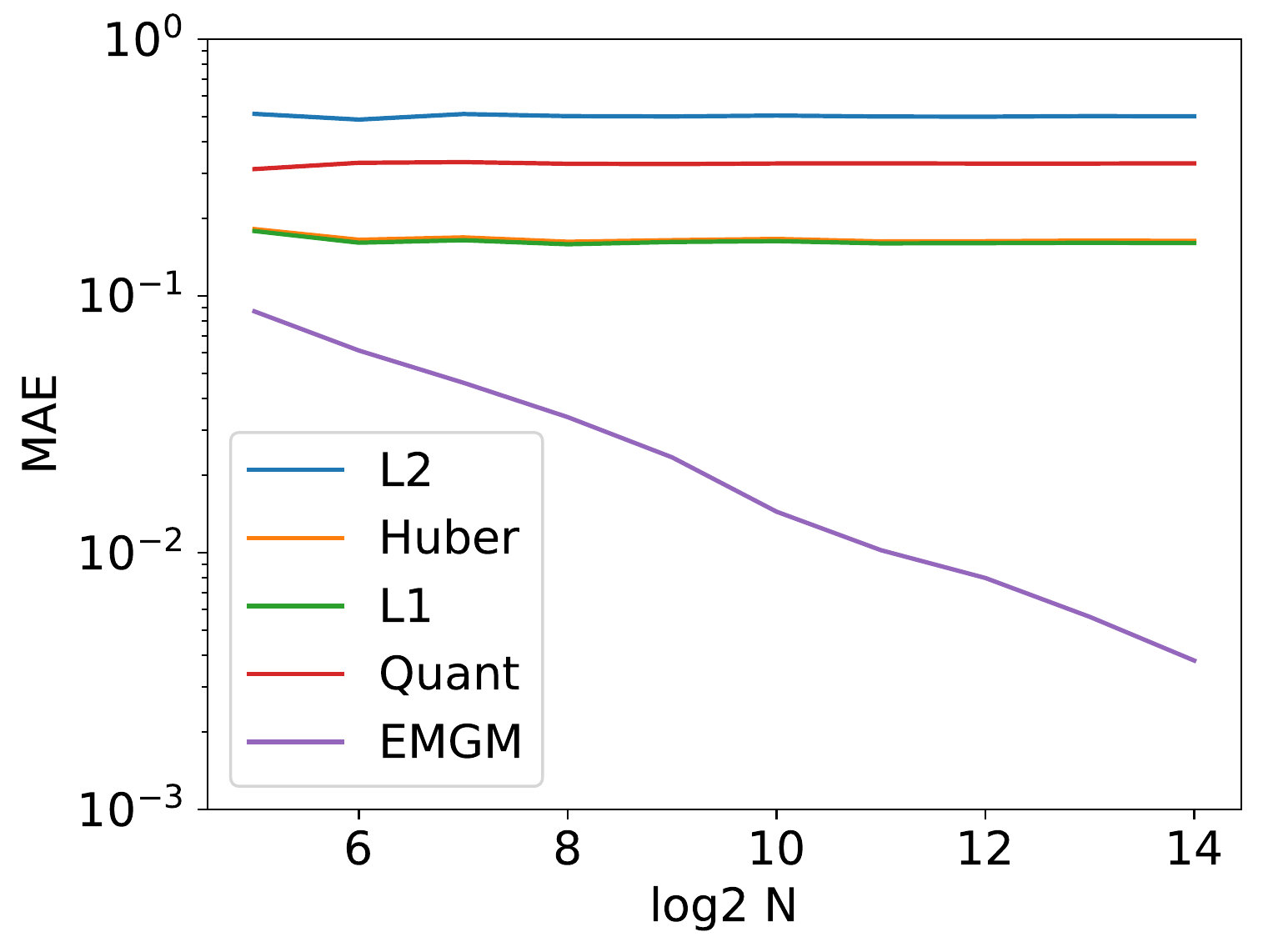}
\includegraphics[width=.49\linewidth]{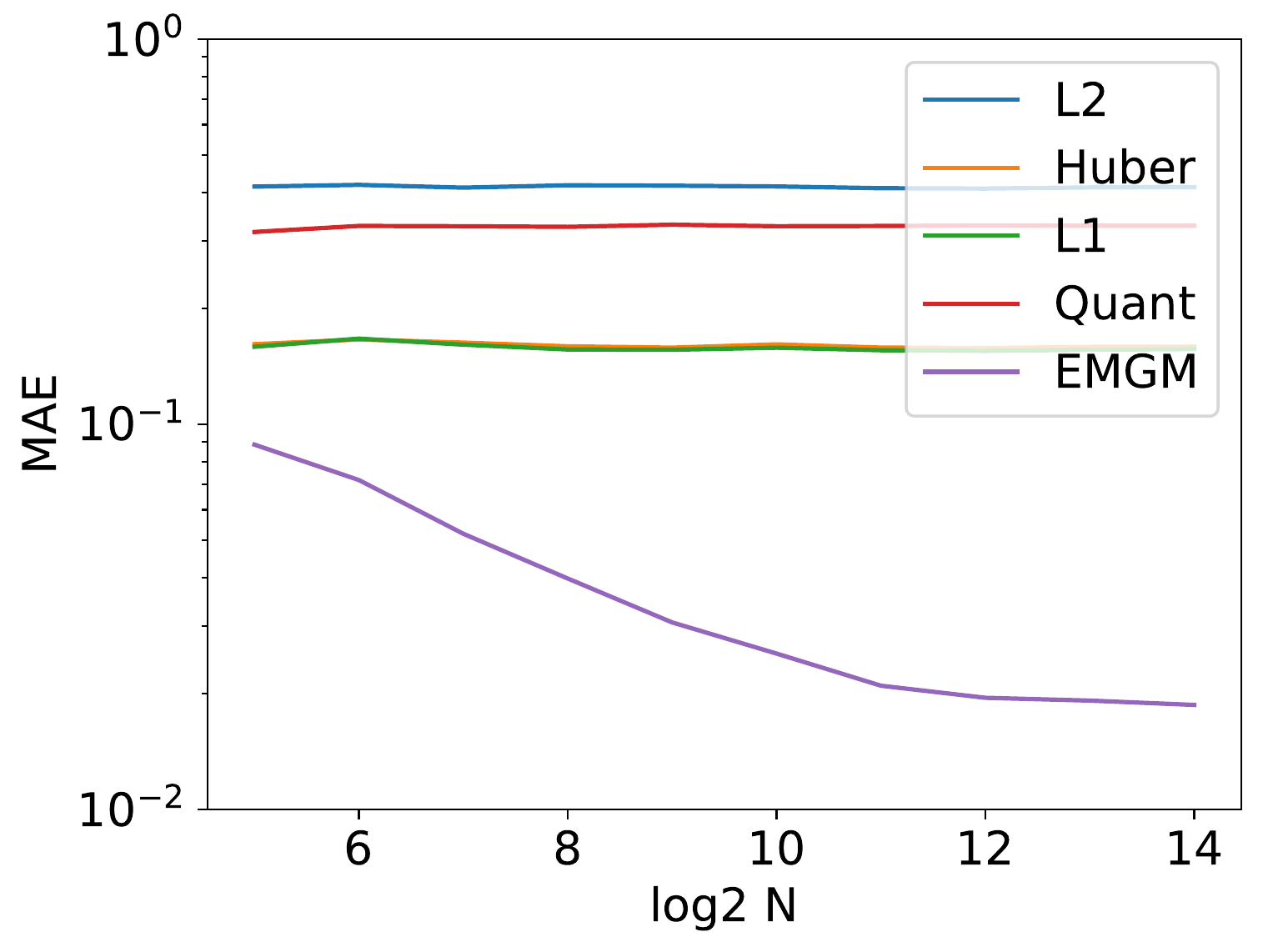}
\caption{ MAE of $b$ as a function of data size $N$.
Left: Exponentially-distributed contaminations. Right: Log-normally-distributed contaminations.
With increasing data size, EMGM exhibits convergent behavior for exponential
contaminations and achieves a low level of error for log-normally distributed contaminations.}
\label{fig_convergenceRegression}
\end{center}
\end{figure}

Figure \ref{fig_convergenceRegression} also highlights this trend.
It depicts the MAE vs data size for 
exponential (left) and log-normal (right) contaminations.
Notably, the left plot shows that the MAE of the EMGM estimate of $b$ decays approximately as $1/\sqrt{N}$.
The other estimates do not exhibit this convergent behavior.
Surprisingly, even if the data is contaminated by log-normal noise, the MAE of EMGM exhibits a strong reduction
with data size until it plateaus at a low error level (see Figure \ref{fig_convergenceRegression} (right) and Table \ref{tab_regressionResults}).

\subsection{Probabilistic Matrix Factorization for Spectroscopy}
\label{sec_PMFForSpectroscopy}

In complex spectroscopic datasets, several types of background signals and systematic errors can contribute to the observed data.
We assume that these unobserved background components
combine linearly with each other and the spectroscopic peaks to form the observed spectrograms.
Therefore, we model the background of the entire dataset B
as a low-rank matrix:
\begin{equation}
\label{eq_backgroundModel}
    B = UV,
\end{equation}
where $U \in \R^{n \times k}$,
$V \in \R^{k \times m}$.
The columns of $U$ can be interpreted as 
the individual background signals,
while the rows of $V$ are the activation of each background signal per spectrogram in the dataset.
Combining the matrix decomposition \eqref{eq_backgroundModel} with the residual model \eqref{eq_likelihood}, we obtain
\begin{equation}
\label{eq_ModelProb}
P( S | U, V, \sigma, \lambda, z) := \prod_{ij} \text{EMGM}_{(UV)_{ij}, \sigma, \lambda, z_{ij} }( S_{ij}).
\end{equation}

$S \in \R^{n \times m}$ is the matrix of measurements, whose $m$ columns consist of spectrograms of length $n$.
Estimating the factors $U, V$ is a type of probabilistic matrix factorization (see Section \ref{sec_ProbMat}).
In the experiments, the factor matrices are optimized with the expectation-maximization algorithm proposed in Section \ref{sec_EMAlgorithm},
with two minor additions:

First, we restrict the columns of $U$ to belong to a Reproducing-Kernel Hilbert Space (RKHS), as background components often exhibit special
characteristics such as smoothness. 
In our experiments, we use an RKHS generated by the RBF kernel, whose length scale is large enough to permit a low-rank factorization of the kernel matrix $K_{ij} = k(x_i, x_j)$.
We calculate the factorization
with an SVD of $K$ as a pre-computation.
Denote by $W$ the left singular vectors 
corresponding to singular values above a specified  precision.
Then $W (W^T U)$ is the projection of the columns of $U$ onto the RKHS.
This matrix-matrix product is $O(nkr)$ where $r$ is the numerical rank of $K$, and $k$ is 
the number of columns of $U$.
The projection occurs in the gradient steps of the maximization step of the EM algorithm, leading to a projected gradient descent algorithm.

Secondly,
we introduce a half-Normal prior on sigma: $\sigma \sim \N_+(0, \sigma_\sigma)$.
This makes the inference of $\sigma$ via the gradient descent algorithm, a generally non-convex problem,
more well-posed and encourages solution with small Gaussian noise variance.
We proceed to estimate $\sigma$ with the algorithm described in Section \ref{sec_EMAlgorithm}.

\subsubsection{Synthetic Spectroscopic Data}
\label{sec_synData}

\begin{figure}
\begin{center}
\includegraphics[width= .95\linewidth]{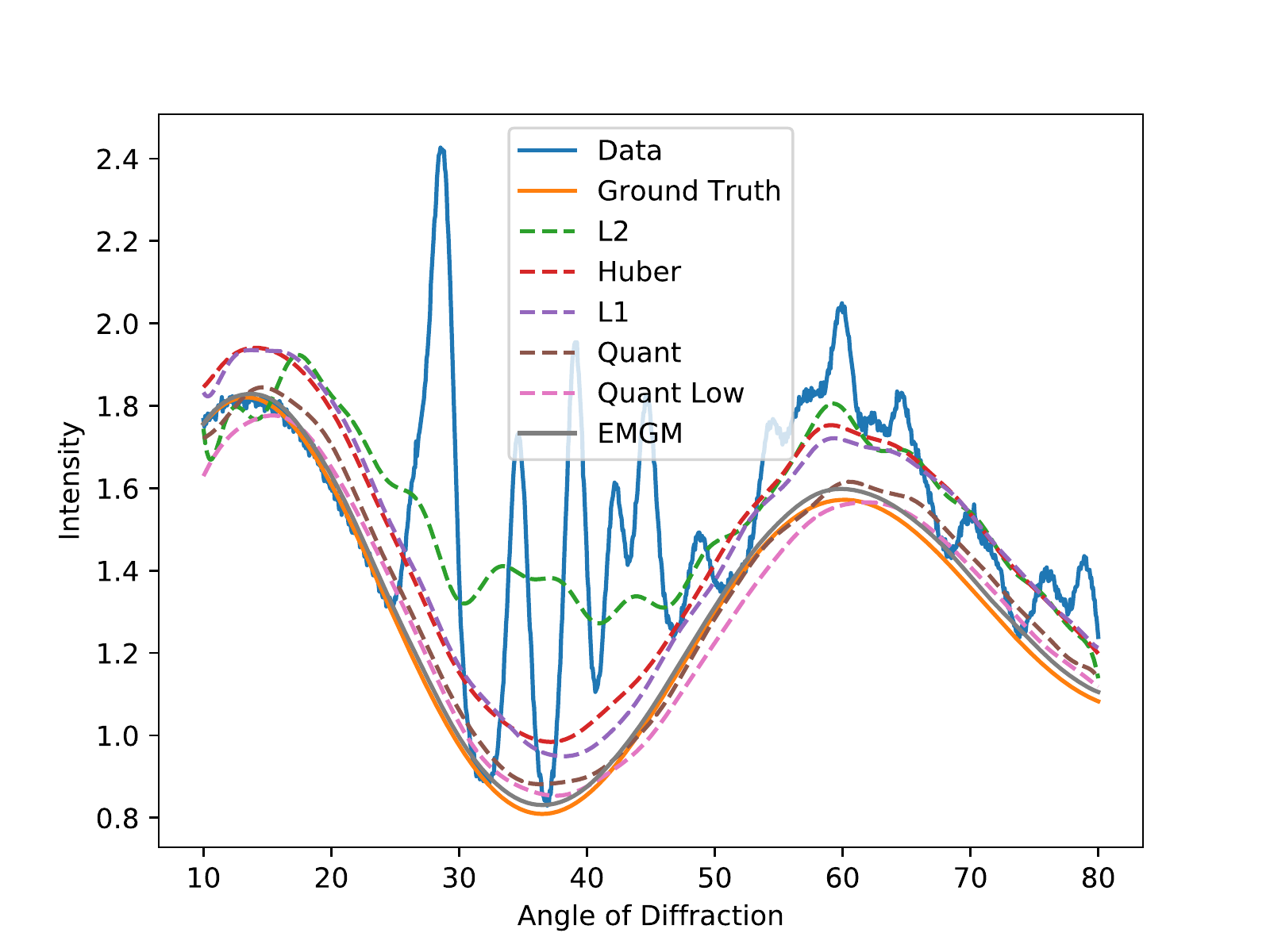}
\caption{Sample background estimates for a data set with 128 synthetically generated X-ray diffraction patterns.
The matrix factorization driven by the EMG mixture closely follows the ground truth, while most other methods overestimate it significantly, thereby absorbing important peaks from the relevant non-background signals.
The estimates of the regression quantile objectives (Quant with $q = .3$, Quant Low with $q = .2$)
are signficantly better though still not as accurate as EMGM.
In contrast to the EMGM, there is no automatic procedure to choose the best parameters for the quantile objective.}
\label{fig_syntheticExample}
\end{center}
\end{figure}

\begin{table}
    \caption{ Error statistics of the spectroscopic background estimation over 32 datasets of $N = 128$ spectrograms with several ranks $k = 1,2,3$.
    EMGM outperforms the other methods in the majority of cases.}
    \label{tab_syntheticErrors}
    \begin{center}
\resizebox{.95\linewidth}{!}{
        \begin{tabular}{ |c|c|c|c|c|c|c| }
    \hline
     $k = 1 $ & $\ell2$ & $\ell1$ & Quant & Quant Low & EMGM  \\ 
        \hline
    mean $\ell2$ & 2.03e-01 & 1.12e-01 & 6.46e-02 & 4.65e-02 & \textbf{3.93e-02} \\
    std $\ell2$ & 1.83e-02 & 1.27e-02 & 8.95e-03 & 7.65e-03 & \textbf{5.34e-03} \\
    mean $\ell1$ & 1.41e-01 & 7.05e-02 & 4.00e-02 & \textbf{2.94e-02} & 3.02e-02 \\
    std $\ell1$ & 1.19e-02 & 7.99e-03 & 5.39e-03 & 4.62e-03 & \textbf{4.11e-03} \\
         \hline
    $k = 2$ & $\ell2$ & $\ell1$ & Quant & Quant Low & EMGM \\ 
        \hline
    mean $\ell2$ & 2.41e-01 & 1.39e-01 & 8.51e-02 & 7.54e-02 & \textbf{6.54e-02} \\
    std $\ell2$ & 2.30e-02 & 2.19e-02 & \textbf{2.15e-02} & 2.57e-02 & 2.88e-02 \\
    mean $\ell1$ & 1.36e-01 & 8.49e-02 & 5.12e-02 & 4.43e-02 & \textbf{4.00e-02} \\
     std $\ell1$ & 1.22e-02 & 1.30e-02 & 1.22e-02 & 1.26e-02 & \textbf{1.09e-02} \\
    \hline
    $k = 3$ & $\ell2$ & $\ell1$ & Quant & Quant Low & EMGM  \\ 
    \hline
    mean $\ell2$ & 2.47e-01 & 1.38e-01 & 8.63e-02 & 7.67e-02 & \textbf{6.05e-02} \\
    std $\ell2$ & 2.88e-02 & 3.06e-02 & 1.81e-02 & \textbf{1.79e-02 }& \textbf{1.79e-02} \\
    mean $\ell1$ & 1.44e-01 & 8.51e-02 & 5.34e-02 & 4.83e-02 & \textbf{3.75e-02} \\
    std $\ell1$ & 1.21e-02 & 1.74e-02 & 1.13e-02 & 1.28e-02 & \textbf{8.57e-03} \\
    \hline
    \end{tabular}
}
\end{center}
\end{table}
We study the behavior of PMF using the EMG mixture 
on a synthetic spectroscopic dataset created using the Materials Project,
an open database which currently contains information for 83,989 inorganic compounds \cite{Jain2013}. 
We randomly selected the spectrograms of a subset of them to generate synthetic data for the background inference task.
In particular, we generated a set of datasets with $N$ spectrograms,
which consist of 1024 datapoints each.
Each spectrogram is a linear combination of an X-ray diffraction pattern, up to three synthetically generated background components ($k = 1,2,3$), and 
Gaussian noise.

We perform the matrix factorization using 5 different objective functions:
$\ell_2, \ell_1$, Huber, quantile, and the EMG mixture.
We compute factorization with the quantile objective function for $q = .3$ and $q = .2$, referred to as Quant and Quant Low, respectively.

Regarding parameter initialization, the length scale of the RBF kernel is $l = 5$.
The factor matrices are set to $U = W(W^T R_U)$, and $V = R_V$ where the elements of $R_U \in \R^{n, k}$, $R_V \in \R^{k, m}$ are drawn from $U(0,1)$.
The rank $k$ is set to its true value in each of the three cases $k = 1,2,3$.
Further, the EMGM's distributional parameters are initialized 
to their maximum-likelihood estimates on
the residuals of the quantile factorization with $q=.3$.

See Figure \ref{fig_syntheticExample} for a visual comparison of the five methods on a sample spectrogram.
The result of the EMG mixture follows the ground truth closely, while
the other objective functions, except for quantile, overestimate the ground truth of the background significantly.

For a more quantitative analysis, see Table \ref{tab_syntheticErrors}.
For the results of each objective function, 
we record the vector $\ell2$ norm and the vector $\ell1$ norm
of the residual matrices.
To summarize these results,
Table \ref{tab_syntheticErrors} shows the average and standard deviation of the error norms over 32 synthetically-generated datasets with $N = 128$ spectrograms.
We report results of the two quantile factorizations in favor of the Huber results, which are comparable to $\ell1$.
The quantile matrix factorization with $q = .2$ is competitive for $k = 1$, especially considering the $\ell 1$ norm.
It is important to note however, that it is a-priori not clear
which quantile will be accurate.
Also, the higher $\ell 2$ norm errors indicate that the estimates are not uniformly accurate, but are unstable across a dataset:
a disadvantage of estimators based on low quantiles.
For $k = 2,3$ the EMGM outperforms all other methods.

\begin{figure}
\begin{center}
\includegraphics[width=.9\linewidth]{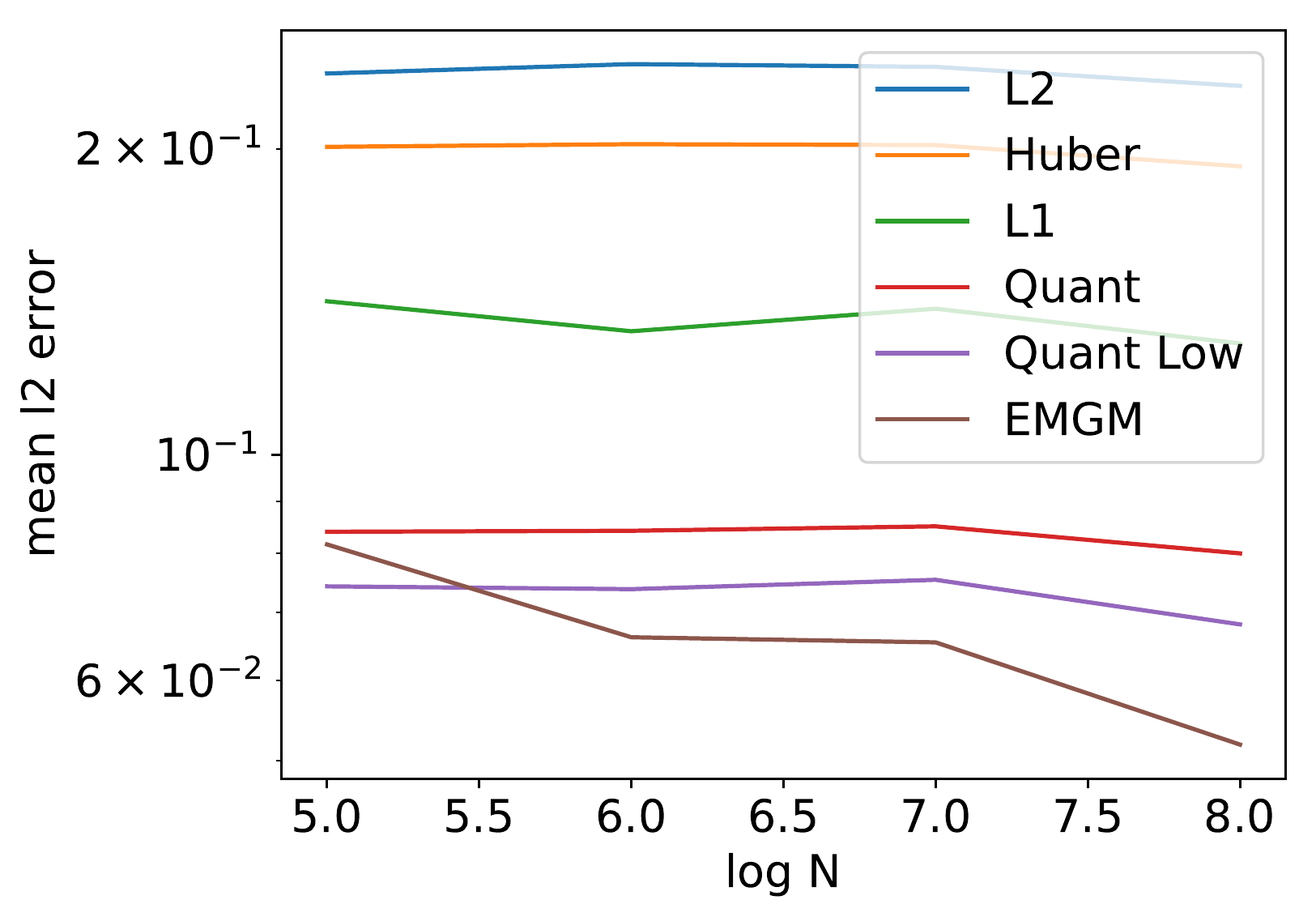}
\caption{ Mean vector $\ell2$ error of the synthetic background estimation task as a function of the number of spectrograms $N$ for $k = 2$.
EMGM improves with increasing data size.}
\label{fig_convergenceFactorization}
\end{center}
\end{figure}

Figure \ref{fig_convergenceFactorization} shows the mean vector $\ell2$ error
of several residual models.
While Quant Low ($q=.2$) starts out very well, it plateaus.
In contrast, the EMGM performs better with increasing amounts of data.

\subsection{
X-Ray Diffraction and Raman Spectroscopy Data}
\label{sec_realWorld}

We illustrate the efficacy of the EMG mixture model to infer the background in a real-world X-ray diffraction (XRD) dataset and a real-world Raman dataset, arising in materials science.
The Raman dataset has 2100 spectrograms each consisting of 1024 points.
The XRD dataset has 186 spectrograms, each consisting of 6400 points.

Prior work on background subtraction in spectroscopy
is mostly based on the application of smoothing operators
on one spectrogram of a dataset at a time.
For example, \cite{NewBckGrnd} uses a cubic spline interpolation on set of heuristically chosen nodes.
\cite{ZhaoRamanBackground} introduced a method called I-ModPoly. It works by iteratively fitting a low-order polynomial to a spectrogram and updating a noise level estimate.
The datapoints above this noise level are ignored for the polynomial fitting.

\begin{figure}
\begin{center}
\includegraphics[width=\linewidth]{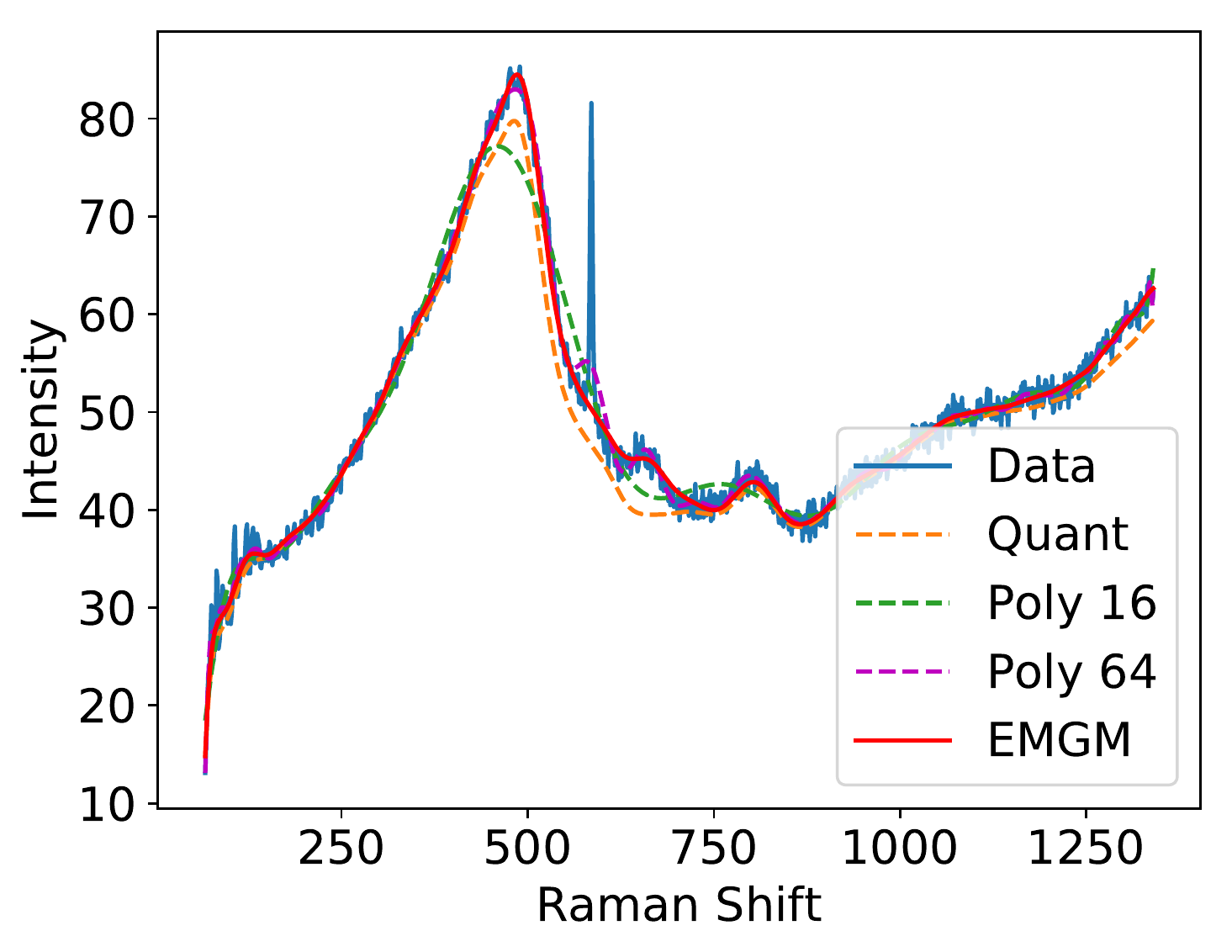}
\includegraphics[width=\linewidth]{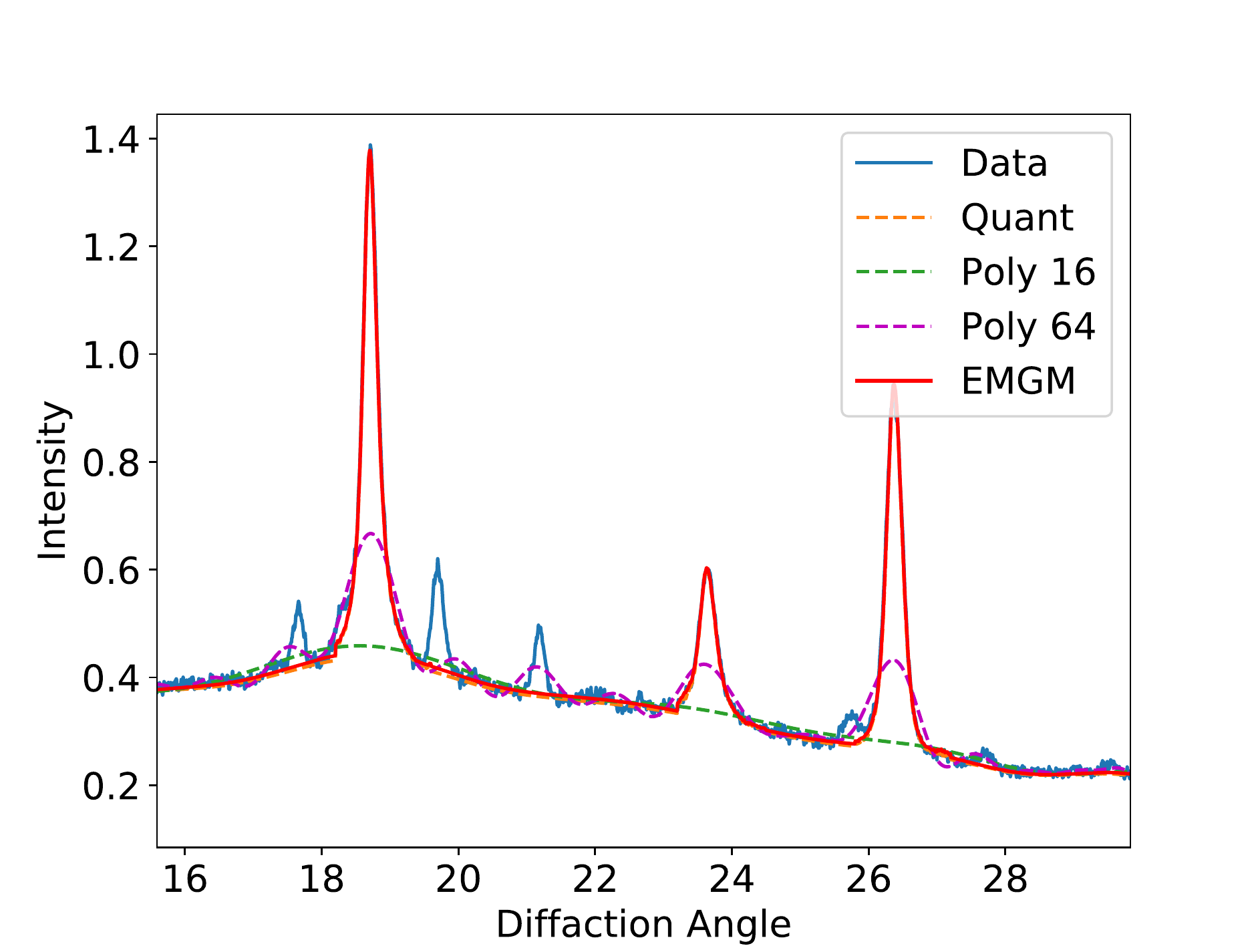}
\caption{\textbf{Top:} A real-world Raman spectrogram with several background models.
Quant ($q = .2$) underestimates the background at 500 - 700 and from 1250 onward.
The polynomial method with low degree also underestimates the background at 500,
while the high degree chips away from the peak at 600.
The EMGM does not suffer from these drawbacks.
\textbf{Bottom:} A real-world XRD spectrogram with substrate peaks at 19, 24, 26.5.
Both EMGM and quantile matrix factorization (with a good, manually chosen $q$ as input) are able to capture the substrate signal
correctly.
The low-order polynomial method overshoots slightly around
the most intense peak at 19.
The high-order polynomial starts to absorb all non-substrate peaks.
}
\label{fig_realComparison}
\end{center}
\end{figure}

In addition to smooth background signals, some datasets are created on a substrate with its own spectroscopic signature,
which is shared among all spectrograms of the dataset.
We consider this spectroscopic signature background signal to be able to subtract it and facilitate the the analysis of the scientifically interesting components of the signal.
Figure \ref{fig_realComparison} (bottom) shows an XRD spectrogram in which the three most intense peaks come from a background source.

Figure \ref{fig_realComparison} (top) shows a sample Raman spectrogram, and
background models calculated by the approach of Section \ref{sec_PMFForSpectroscopy} with quantile factorization ($q = .2$) and EMGM, and
the method of \cite{ZhaoRamanBackground}
with two different polynomial degrees.
The EMGM result is the only one that does not exhibit drawbacks (see caption for details).

Figure \ref{fig_realComparison} (bottom) shows
a sample XRD spectrogram with background models calculated by the same methods as above.
Interestingly, the EMGM and quantile methods both
infer the substrate signature and background signals correctly, as verified by human experts.
As we have seen in Section \ref{sec_synData}, 
quantile factorization can work well if the quantile is chosen appropriately, 
so this result does not come as a complete surprise.

\section{Conclusion}

We introduced the exponentially-modified Gaussian (EMG) mixture residual model, which is well suited to model residuals that are contaminated by a distribution with positive support.
We proved two convexity results for the negative logarithm of the EMG density, 
and further introduced an expectation-maximization algorithm
for optimizing the EMG mixture model.
We compared the EMG mixture against commonly-used residual models such as $\ell_1$, the Huber loss, and regression quantiles,
showing its convergence for exponentially-distributed contaminations.
We incorporated the EMG mixture into a probabilistic matrix factorization framework, motivated by applications in spectroscopy.
We showed how this approach is effective in inferring background signals
and systematic errors in data arising from X-ray  diffraction and Raman spectroscopy, dramatically outperforming existing approaches and revealing the data's physically meaningful components.

We hope that our work will inspire other researchers to pursue possible extensions.
For example, while our real-world data comes from materials science,
the methods should be widely applicable to spectroscopic data
arising in other scientific domains (e.g. astronomy, physics, biology) and also
spectroscopic data arising in other domains (e.g. music, speech, animal vocalizations).

\pagebreak

\bibliography{references.bib}

\begin{thebibliography}{18}
\providecommand{\natexlab}[1]{#1}
\providecommand{\url}[1]{\texttt{#1}}
\expandafter\ifx\csname urlstyle\endcsname\relax
  \providecommand{\doi}[1]{doi: #1}\else
  \providecommand{\doi}{doi: \begingroup \urlstyle{rm}\Url}\fi

\bibitem[Bertsekas(2008)]{Bertsekas2008}
Bertsekas, D.~P.
\newblock \emph{{Nonlinear programming}}.
\newblock Athena Scientific, 2nd edition, September 2008.
\newblock ISBN 1886529000.

\bibitem[Carr et~al.(2009)Carr, Madan, and H~Smith]{Carr2008SaddlepointMF}
Carr, P., Madan, D., and H~Smith, R.
\newblock Saddle point methods for option pricing.
\newblock \emph{The Journal of Computational Finance}, 13:\penalty0 49--61, 09
  2009.
\newblock \doi{10.21314/JCF.2009.198}.

\bibitem[Dempster et~al.(1977)Dempster, Laird, and Rubin]{Dempster19977}
Dempster, A.~P., Laird, N.~M., and Rubin, D.~B.
\newblock Maximum likelihood from incomplete data via the em algorithm.
\newblock \emph{Journal of the Royal Statistical Society. Series B
  (Methodological)}, 39\penalty0 (1):\penalty0 1--38, 1977.
\newblock ISSN 00359246.
\newblock URL \url{http://www.jstor.org/stable/2984875}.

\bibitem[Fujisawa \& Eguchi(2008)Fujisawa and Eguchi]{Fujisawa2008}
Fujisawa, H. and Eguchi, S.
\newblock Robust parameter estimation with a small bias against heavy
  contamination.
\newblock \emph{Journal of Multivariate Analysis}, 99\penalty0 (9):\penalty0
  2053 -- 2081, 2008.
\newblock ISSN 0047-259X.
\newblock \doi{https://doi.org/10.1016/j.jmva.2008.02.004}.
\newblock URL
  \url{http://www.sciencedirect.com/science/article/pii/S0047259X08000456}.

\bibitem[Golubev(2010)]{EMGBio}
Golubev, A.
\newblock Exponentially modified gaussian (emg) relevance to distributions
  related to cell proliferation and differentiation.
\newblock \emph{Journal of Theoretical Biology}, 262\penalty0 (2):\penalty0 257
  -- 266, 2010.
\newblock ISSN 0022-5193.

\bibitem[Green et~al.(2013)Green, Takeuchi, and Hattrick-Simpers]{Green2013}
Green, M.~L., Takeuchi, I., and Hattrick-Simpers, J.~R.
\newblock {Applications of high throughput (combinatorial) methodologies to
  electronic, magnetic, optical, and energy-related materials}.
\newblock \emph{Journal of Applied Physics}, 113, 2013.

\bibitem[Huber(1964)]{Huber1964}
Huber, P.~J.
\newblock Robust estimation of a location parameter.
\newblock \emph{The Annals of Mathematical Statistics}, 35\penalty0
  (1):\penalty0 73--101, 1964.
\newblock ISSN 00034851.
\newblock URL \url{http://www.jstor.org/stable/2238020}.

\bibitem[Jain et~al.(2013)Jain, Ong, Hautier, Chen, Richards, Dacek, Cholia,
  Gunter, Skinner, Ceder, and Persson]{Jain2013}
Jain, A., Ong, S.~P., Hautier, G., Chen, W., Richards, W.~D., Dacek, S.,
  Cholia, S., Gunter, D., Skinner, D., Ceder, G., and Persson, K.~a.
\newblock {The Materials Project: A materials genome approach to accelerating
  materials innovation}.
\newblock \emph{APL Materials}, 1\penalty0 (1):\penalty0 011002, 2013.
\newblock ISSN 2166532X.
\newblock \doi{10.1063/1.4812323}.
\newblock URL \url{http://link.aip.org/link/AMPADS/v1/i1/p011002/s1\&Agg=doi}.

\bibitem[Kanamori \& Fujisawa(2015)Kanamori and Fujisawa]{Kanamori2015}
Kanamori, T. and Fujisawa, H.
\newblock Robust estimation under heavy contamination using unnormalized
  models.
\newblock \emph{Biometrika}, 102\penalty0 (3):\penalty0 559--572, 2015.
\newblock \doi{10.1093/biomet/asv014}.
\newblock URL \url{http://dx.doi.org/10.1093/biomet/asv014}.

\bibitem[Koenker \& Bassett(1978)Koenker and Bassett]{Bassett1978}
Koenker, R. and Bassett, G.
\newblock Regression quantiles.
\newblock \emph{Econometrica}, 46\penalty0 (1):\penalty0 33--50, 1978.
\newblock ISSN 00129682, 14680262.
\newblock URL \url{http://www.jstor.org/stable/1913643}.

\bibitem[Koenker \& Hallock(2001)Koenker and Hallock]{koenker2001quantile}
Koenker, R. and Hallock, K.~F.
\newblock Quantile regression.
\newblock \emph{Journal of economic perspectives}, 15\penalty0 (4):\penalty0
  143--156, 2001.

\bibitem[Mnih \& Salakhutdinov(2008)Mnih and Salakhutdinov]{ProbMF}
Mnih, A. and Salakhutdinov, R.~R.
\newblock Probabilistic matrix factorization.
\newblock In \emph{Advances in neural information processing systems}, pp.\
  1257--1264, 2008.

\bibitem[Neal \& Hinton(1999)Neal and Hinton]{Neal:1999:VEA:308574.308679}
Neal, R.~M. and Hinton, G.~E.
\newblock Learning in graphical models.
\newblock chapter A View of the EM Algorithm That Justifies Incremental,
  Sparse, and Other Variants, pp.\  355--368. MIT Press, Cambridge, MA, USA,
  1999.
\newblock ISBN 0-262-60032-3.
\newblock URL \url{http://dl.acm.org/citation.cfm?id=308574.308679}.

\bibitem[Palmer et~al.(2011)Palmer, Horowitz, Torralba, and Wolfe]{EMGPsych}
Palmer, E.~M., Horowitz, T.~S., Torralba, A., and Wolfe, J.~M.
\newblock What are the shapes of response time distributions in visual search?
\newblock \emph{Journal of Experimental Psychology: Human Perception and
  Performance}, 37\penalty0 (1):\penalty0 58--71, 2011.

\bibitem[Takeuchi et~al.(2002)Takeuchi, Bengio, and Kanamori]{Takeuchi2002}
Takeuchi, I., Bengio, Y., and Kanamori, T.
\newblock Robust regression with asymmetric heavy-tail noise distributions.
\newblock \emph{Neural Computation}, 14\penalty0 (10):\penalty0 2469--2496,
  2002.
\newblock \doi{10.1162/08997660260293300}.
\newblock URL \url{https://doi.org/10.1162/08997660260293300}.

\bibitem[Virtanen(2007)]{Virtanen2007}
Virtanen, T.
\newblock Monaural sound source separation by nonnegative matrix factorization
  with temporal continuity and sparseness criteria.
\newblock \emph{IEEE Transactions on Audio, Speech, and Language Processing},
  15\penalty0 (3):\penalty0 1066--1074, March 2007.
\newblock ISSN 1558-7916.
\newblock \doi{10.1109/TASL.2006.885253}.

\bibitem[Yi et~al.(2015)Yi, Liu, Wang, Chen, Peng, Zhao, He, and
  Zhao]{NewBckGrnd}
Yi, L., Liu, Z., Wang, K., Chen, M., Peng, S., Zhao, W., He, J., and Zhao, G.
\newblock A new background subtraction method for energy dispersive x-ray
  fluorescence spectra using a cubic spline interpolation.
\newblock \emph{Nuclear Instruments and Methods in Physics Research Section A:
  Accelerators, Spectrometers, Detectors and Associated Equipment},
  775\penalty0 (Supplement C):\penalty0 12 -- 14, 2015.
\newblock ISSN 0168-9002.

\bibitem[Zhao et~al.(2007)Zhao, Lui, McLean, and Zeng]{ZhaoRamanBackground}
Zhao, J., Lui, H., McLean, D.~I., and Zeng, H.
\newblock Automated autofluorescence background subtraction algorithm for
  biomedical raman spectroscopy.
\newblock \emph{Appl. Spectrosc.}, 61\penalty0 (11):\penalty0 1225--1232, 11
  2007.

\end{thebibliography}
\bibliographystyle{icml2019}

\end{document}